\documentclass{article}

\usepackage[final,nonatbib]{neurips_2019}

\usepackage{hyperref}       
\usepackage{url}            
\usepackage{booktabs}       
\usepackage{amsfonts}       
\usepackage{nicefrac}       
\usepackage{microtype} 
\usepackage{epstopdf}
\usepackage{rotating,booktabs,multirow}
\usepackage{graphicx}
\usepackage{amsmath,amssymb} 
\usepackage{color}
\usepackage{subcaption}
\usepackage{wrapfig,lipsum}

\usepackage{algorithm}
\usepackage{algorithmic}
\usepackage[algo2e]{algorithm2e}
\usepackage{rotating,booktabs,multirow}
\usepackage{xparse}
\usepackage{mathtools}
\usepackage{amsfonts}
\usepackage{caption}
\usepackage[numbers]{natbib}

\title{Boosting Mapping Functionality of Neural Networks via Latent Feature Generation based on Reversible Learning}

\author{
  Jongmin Yu\thanks{\url{https://sites.google.com/view/jongmin-yu-cv/home}}\\
  Department of Electrical Engineering, Curtin University, Perth, WA 15213, Australia\\
  School of Electrical Engineering and Computer Science, GIST, Gwangju, 61005, Korea (Republic of)\\
  \texttt{jm.andrew.yu@gmail.com} \\
}

\begin{document}

\maketitle

\begin{abstract}
This paper addresses a boosting method for mapping functionality of neural networks in visual recognition such as image classification and face recognition. We present reversible learning for generating and learning latent features using the network itself.  By generating latent features corresponding to hard samples and applying the generated features in a training stage, reversible learning can improve a mapping functionality without additional data augmentation or handling the bias of dataset. We demonstrate an efficiency of the proposed method on the MNIST, Cifar-10/100, and Extremely Biased and poorly categorized dataset (EBPC dataset). The experimental results show that the proposed method can outperform existing state-of-the-art methods in visual recognition. Extensive analysis shows that our method can efficiently improve the mapping capability of a network.
\end{abstract}

\section{Introduction}
In visual recognition studies using neural networks, such as image classification \cite{lecun1998gradient,he2016deep} and face recognition \cite{wen2016discriminative,liu2017sphereface}, the networks can be thought a mapping function between high dimensional data and the low dimensional space. Commonly used approaches to improving mapping functionality and recognition accuracies in such visual recognition studies, are modifying network connections \cite{he2016deep,huang2017densely} or adjusting loss functions \cite{liu2017sphereface,wen2016discriminative}. Above methods show outstanding performances when the well-classified and balanced datasets are provided.

There is a possibility that a model has poor mapping capability due to extrinsic factors related to the uncertainty of a given dataset e.g., class-wise unbalance and noisy label. These issues can sometimes be addressed by handling hard samples. A sample is considered as a hard sample when it is on the wrong side of the correct decision boundary \cite{viola2001robust} or in the margin of the hyperplanes for classification \cite{dollar2014fast}. Hard samples are frequently observed when an unbalanced and not-well classified datasets are given, because existing learning methods usually have to inductive bias toward the dominant classes if training data are unbalanced, resulting in poor minority class recognition performance \cite{dong2018imbalanced}. Most approaches to solving this issues are dataset resampling \cite{chawla2002smote} setting the balanced proportion of label and data to train a network model, and applying weight decided by the training loss using hard sampling mining \cite{kumar2010self, chang2017active, jiang2017mentornet}. Recently, a hard sample generation methods based on deep neural networks has been proposed to tackle the hard sample problem caused from dataset imbalance \cite{schroff2015facenet, wu2017sampling}.

However, without an appropriate definition for the bias of dataset, handling the dataset bias is inherently ill-defined \cite{ren2018learning}. Moreover, using these methods without a proper definition of dataset bias may lead to poorly optimized mapping results when training networks. Also, hard samples can appear even if neural networks are trained with a physically well-balanced and correctly classified dataset. Consequently, it is necessary to develop a method for boosting a mapping functionality of neural networks which is invariant to the bias of datasets and does not require meta information such as the definition of dataset bias and sample proportions between classes.

In this work, we present a reversible network learning (RNL) which can allow the reversiblity to neural networks, and we propose a generation and learning of hard-sample corresponding to latent features based on RNL. The reversible network (RN) is inspired by the auto-encoder. However, it differs than AE since it focuses on reconstructing, generating, and learning latent features to improve supervised learning performance. In learning the RNs, the network can generate the latent features regarded as samples which have a lower likelihood, and apply the features to network learning. We demonstrate an efficiency of the proposed method using image classification problems. The experimental results show that the networks trained by the proposed manner outperform the others.

The key contributions of this paper are summarized as three points: First, we propose a RN for generation and learning of latent features related to hard samples; Second, the proposed method is easily applied to the various network structures and loss functions, and the resulting models perform substantially better than existing ones. Third, we provide extensive experimental results, including the comparison for recognition performances between the proposed methods and other models and the relation between latent features and hard samples.

This paper is organized as follows. We describe the RN and how to generate and learn the hard sample corresponding latent features using the network in Section. 2. In Section 3, we provide experimental results to demonstrate the efficiency of the proposed method. We conclude this paper in Section. 5.

\section{Reversible Network Learning}
\subsection{Reversibility of Neural Network}
In supervised learning manner, neural network conduct feed-forward process. Given the input sample $x$, the networks map the samples $x$ into latent space and extract latent feature $\alpha$, and compute the confidence value related to each class. The primary goal of the learning networks is minimizing errors such as classification error. The reversibility of a neural network is not considered as a significant issue usually.

In a $n$-class classification problem setting using a neural network, suppose the network consists of two functions: an encoding function $f:x\rightarrow{}\alpha$, where $x$ and $\alpha$ are an input sample and the corresponding latent feature, and probabilistic model $p(\alpha|\theta_{i})$ to compute the likelihood of $\alpha$ corresponding to class $i$ for each input sample using given $\alpha$ and the model parameter $\theta_{i}$. The entire process can be represented by $p\circ{}f(X)=p(f(x)|\theta_{i})$, and the predicted class $c$ is decided as follows: $c=\text{argmax}_{i}p(f(x)|\theta_{i})$. 

Under the above setting, the neural network is reversible if the network satisfies the following condition from a given sample $x$:
\begin{equation}
\label{equ:1}
\begin{aligned}
x\equiv(p\circ{}f)^{-1})(p\circ{}f(x)).
\end{aligned}
\end{equation}
However, since the network components such as activation functions and connectivity are sometimes non-invertible, it is difficult to build a mathematically RN in practice \cite{xu2014deep}. For instance, the softmax function is non-linear, and the inverse form of softmax function $\mathcal{S}$ is regarded as: $\mathcal{S}^{-1}(o_{i}) = \text{ln}(o_{i})+c$, where $o_{i}$ is the $i$ class output of softmax function corresponding the latent feature $\alpha_{i}$, and $c$ denotes a constant. since originally $c$ represents $\text{ln}\sum_{j}e^{\alpha^{j}}$, and this replacement make difficulty when neural networks take reversibility. 

\begin{figure}
	\vspace{-0.2cm}
	\begin{center}
		\centerline{\includegraphics[width=\textwidth]{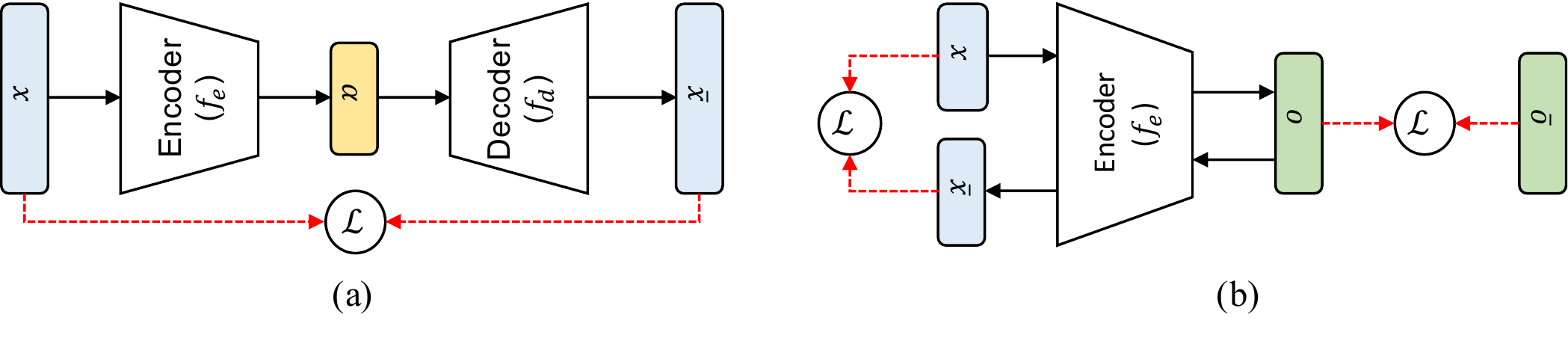}}
		\caption{(a) denotes the workflow of an autoencoder (AE). (b) represents the workflow of the reversible network (RN). $x$ is a given sample, and $\bar{x}$ denotes the reconstruction of $x$ on AE and RN. $\alpha$ of (a) represents the latent features of AE. $o$ and $\bar{o}$ defines the network outputs and corresponding labels. Black solid lines show the working process of each model. Red dotted lines denote loss functions and assigned features for the functions.}
		\label{fig:1}
	\end{center}
	\vspace{-0.2cm}
\end{figure}

Despite the neural network is theoretically irreversible. We can develop a learning-based approach to improve the reversibility of a neural network using the condition in Eq. \ref{equ:1} and reconstruction manner of AE.

To improve the reversibility of neural networks in supervised learning manner, RN conducts two processes: feed-forward and feed-backwards processes. The feed-forward process is a general process of supervised learning. In classification problem setting, the feed-forward process generates an output of networks for classification: $p\circ{}f(X)=p(f(x)|\theta_{i})$. In a classification problem setting, the goal of the feed-forward process is computing the likelihood related to each class, and the optimization scheme conducts to minimize a classification error. The cross-entropy with softmax function is commonly used as a cost function in classification problem setting defined as follows:
\begin{equation}
\label{equ:2}
\begin{aligned}
\mathcal{L}_{ff}(o,\bar{o})=-\sum^{C}_{i=1}\bar{o}_{i}\text{log}(o_{i}),\;  o_{i}=\frac{e^{\alpha_{i}}}{\sum^{C}_{j=1}e^{\alpha_{j}}},
\end{aligned}
\end{equation}
where $C$ is the dimensionality of a final layer, and it is usually equal to the number of classes. $o$ and $\bar{o}$ are an output of the feed-forward process of a network model and the corresponding label. The network output $o$ is decided by computing the softmax function with latent feature $\alpha$, and it can be interpreted as a likelihood for each class of a given sample $x$. 

On the other hands, the feed-backwards process reconstructs the input samples thought the reverse process represented by:
\begin{equation}
\label{equ:3}
\begin{aligned}
(p\cdot{}f)^{-1}:o\rightarrow{}\bar{x}, 
\end{aligned}
\end{equation}
where $\bar{x}$ is the reconstruction results from a given value $o$. Unfortunately, as mentioned above, because of some network components such as non-linear activation functions and irreversible network connections, it is difficult to make a mathematically exact inverse network. 

In RN, the feed-backwards process conducts a reverse process inspired by AE. The reverse process for fully connected networks defined by:
\begin{equation}
\label{equ:4}
\begin{aligned}
\gamma(o\cdot{}W_{t}^{T}+b_{t-1})=\bar{\alpha},
\end{aligned}
\end{equation}
where $W_{t}^{T}$ is the transpose matrix of a weight matrix $W$ in $t^{th}$ fully connected layer, and $b_{t-1}$ is the biase of the previous layer of the $t^{th}$ layer. $\gamma$ is an activation function such as rectified linear unit \cite{}, softmax function, and hyperbolic tangent function. In Equation. 4, $W_{t}^{T}\cdot{}o$ can be represented as follows:
\begin{equation}
\begin{aligned}
\begin{bmatrix}
    o_{1} & \dots  & o_{m}
\end{bmatrix}
\cdot{}
\begin{bmatrix}
    w_{11} & \dots  & w_{1n} \\
    w_{21} & \dots  & w_{2n} \\
    \vdots &  \ddots & \vdots \\
    w_{m1} &  \dots & w_{mn}
\end{bmatrix}^{T}
=
\begin{bmatrix}
    o_{1} & \dots  & o_{m}
\end{bmatrix}
\cdot{}
\begin{bmatrix}
    w_{1}^{T} \\
    w_{2}^{T} \\
    \vdots  \\
    w_{m}^{T} 
\end{bmatrix}
=\sum^{n}_{i=0}o\cdot{}w_{1}^{T},
\end{aligned}
\end{equation}
where $w_{ij}$ is element of the $(i,j)$ coordinate in the weight matrix $W$, and $w_{i}$ is the column vector of $W$. $o$ and $o_{i}$ are the output and $i$-th element of the output. Additionally, the reverse process for convolutional layer is replaced by the deconvolutional layer \cite{xu2014deep}.

\begin{wrapfigure}{R}{0.35\textwidth}
\centering
\includegraphics[width=0.34\textwidth]{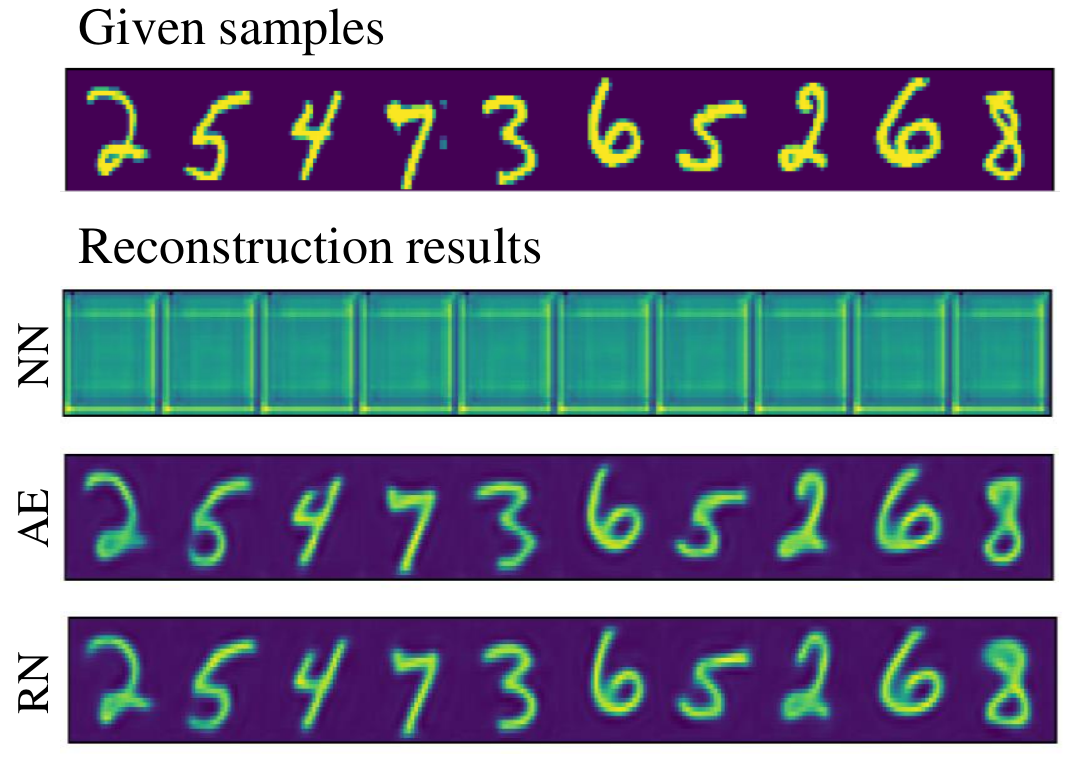}x
\caption{Example samples of the reconstruction results on general neural network (NN), autoencoder (AE), and reversible network (RN). The visualisation results show ordinarily trained network cannot ensure the reversibility.}
	\vspace{0.2cm}
\label{fig:2}
\end{wrapfigure}

Above feed-backward process $(p\circ{}f)^{-1}$ products the reconstruction results $\bar{x}$, and the reconstruction results are applied to maximize the reversibility of neural networks by minimizing the reconstruction error based on mean square error as follows:
\begin{equation}
\label{equ:5}
\begin{aligned}
\mathcal{L}_{fb}(x,\bar{x})=\sum^{W}_{i=0}\sum^{H}_{j=0}|x_{ij}-\bar{x}_{ij}|^{2},
\end{aligned}
\end{equation}
where $x_{ij}$ and $\bar{x}_{ij}$ is $(i,j)$ elements of input $x$ and feed-backwarding results $\bar{x}$. The reversibility maximization via minimizing reconstruction error is inspired from AE. 

However, RN and AE have different objectives methodologically. AE is one of unsupervised learning manners, and the goal of AE is to minimize a reconstruction error between an input sample and a reconstructed result, and this process does not consider classes of samples. This helps AE to learn significant representations from given samples. On the other hands, the goal of training RN is both minimize a recognition accuracy and a reconstruction error. This can be regarded as a class-wise embedding of latent features depending on specific classes since the learning of RN includes clustering process of the latent features based on their likelihoods by conducting two minimizations for the recognition accuracy and the reconstruction error simultaneously. This property of RN may help to reconstruct latent features corresponding to hard samples. Figure \ref{fig:1} illustrates the workflows and the methodological difference between RN learning and an AE. To apply the above two objectiveness to train the network, we used straightforward aggregation to compute to the total loss function. We aggregate classification loss on the feed-forward process and reconstruction loss on the feed-backwards process as follows:
\begin{equation}
\label{equ:6}
\begin{aligned}
\mathcal{L}(x,o) = \mathcal{L}_{ff}(o,\bar{o})+\mathcal{L}_{fb}(x,\bar{x})
\end{aligned}
\end{equation}
In our experiment, other non-differential operations including pooling or another downsampling are replaced to an upsampling function based on simple image transformation methods. Figure \ref{fig:2} shows the reconstruction results of latent features using normal classification network (a.k.a., neural network), AE, and the RN. Parameter setting~. As shown in Figure \ref{fig:2}, the network trained by classification loss only shows poor reconstruction results compared to the others. 

\begin{figure}
        \centering
        \begin{subfigure}{0.33\textwidth}
            \includegraphics[width=\textwidth]{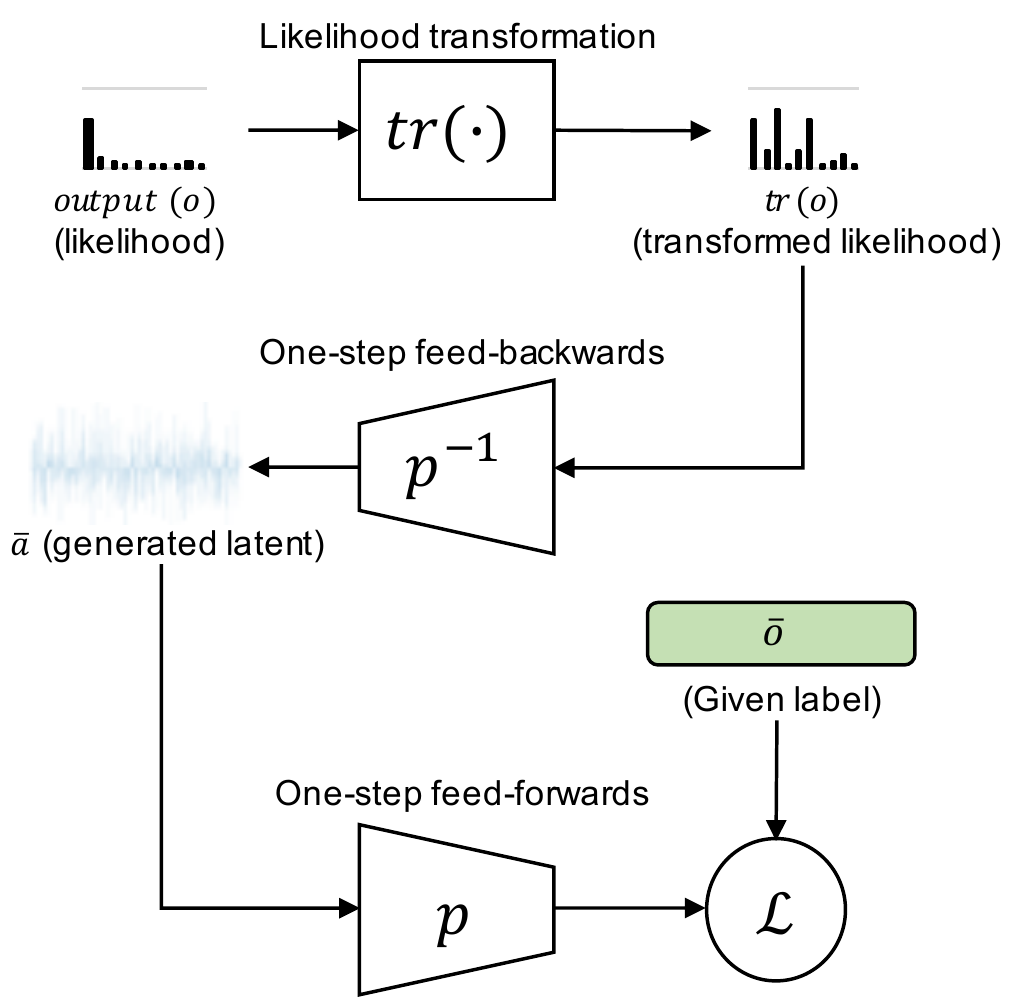}
	\caption{}
    \label{fig:3:a}
        \end{subfigure}
        \hfill
        \begin{subfigure}{0.65\textwidth}
            \includegraphics[width=\textwidth]{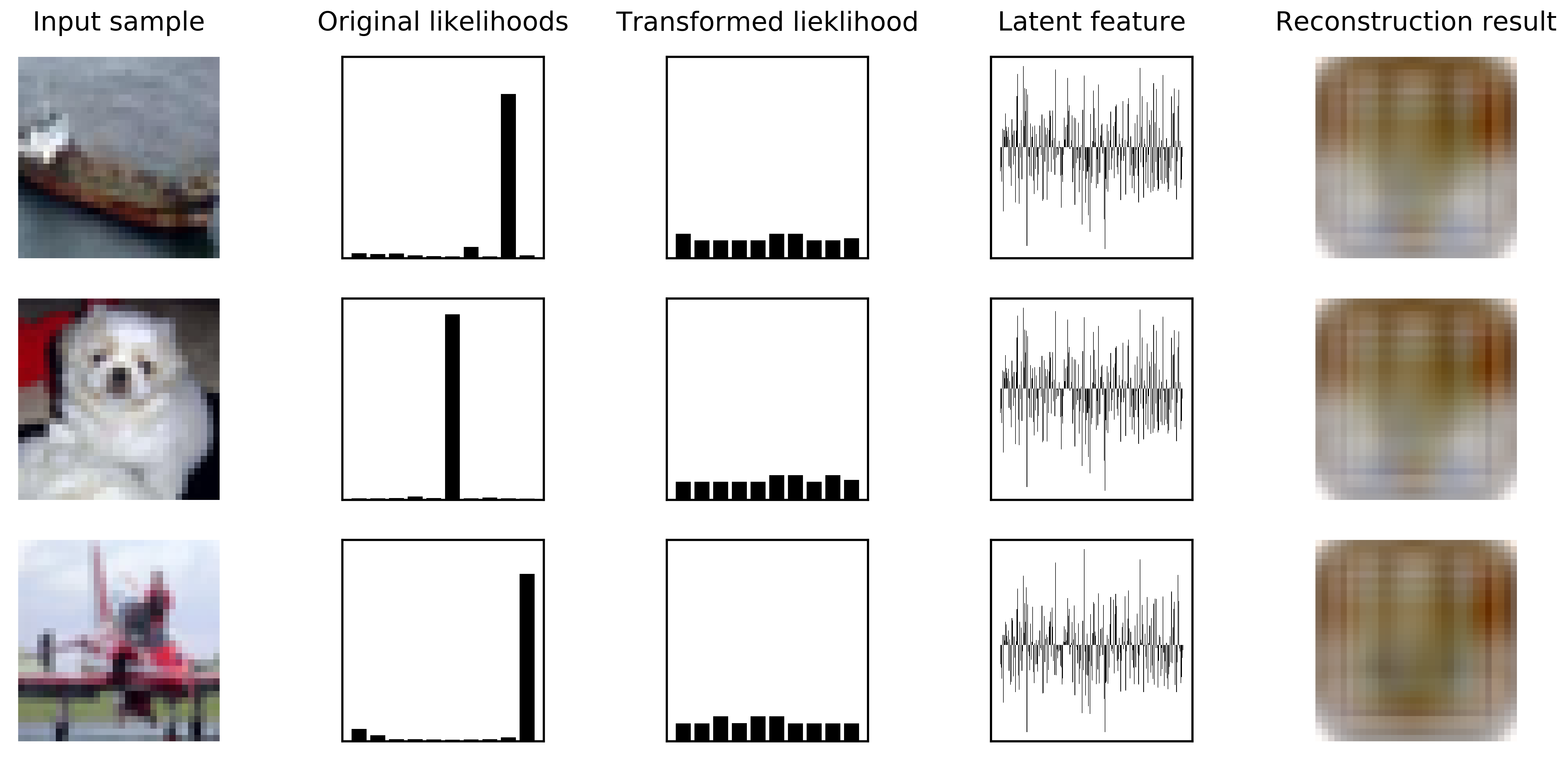}
	\caption{}
    \label{fig:3:b}
        \end{subfigure}
        \hfill
        \caption{(a) illustrates the methodology for generating the latent features corresponding to hard samples. (b) shows the visualization results using Cifar10 dataset. Each column represents input samples, original likelihood (a.k.a., network outputs), transformed likelihoods, generated latent features, and the reconstruction results of the features.}
\label{fig:3}  
\vspace{-1ex}
\end{figure}

\subsection{Latent Feature Generation and Learning}
Inherently, the most simple approach to boost the mapping functionality of neural networks is providing a large-scale and well-categorized dataset which can be used to train the various variations of each class. However, it is difficult to construct the dataset in practice. When neural networks train the biased dataset, the learned features are biased to the dominant samples, and the other samples which are not included in the dominant sample set, are considered as hard samples and it can be classified into wrong class in the test phases.

\begin{algorithm}
	\SetAlgoLined
	\KwIn{Input sample $(x,\bar{o})$, where $x$ and $\bar{o}$ are an input data and corresponding label.}
	\KwResult{The optimized network parameters $\boldsymbol{\theta}=\{\boldsymbol{W},\boldsymbol{b}\}$, where $\boldsymbol{W}$ and $\boldsymbol{b}$ are the sets of weight and bias parameters of the network model.}
	\For{The number samples in a batch}{
        $\cdot$ \textbf{The feed-forward} \\Compute the network output: $(p\circ{}f)(x)$=$o$\\
        $\cdot$ \textbf{The feed-backward} \\Reconstruct the input sample using the network output $o$: $(p\circ{}f)^{-1}(o)=\bar{x}$\\
        $\cdot$ \textbf{The latent feature generation} \\Generate the latent feature corresponding to hard samples: $p^{-1}(\text{tr}(o)|\theta)=\bar{\alpha}$\\ 
	$\cdot$\textbf{One-step feed-forward} \\Compute the output using the generated features: $p(\bar{\alpha}|\theta)=\hat{o}$\\ 
        $\cdot$ \textbf{Loss computing}\\ $\mathcal{L}(x,o)=\mathcal{L}_{ff}(o,\hat{o},\bar{o})+\mathcal{L}_{fb}(x,\bar{x})$\\
    	$\cdot$ \textbf{Update parameters}\\ $\theta = \theta+\gamma\frac{\delta\mathcal{L}}{\delta\theta}$, where $\gamma$ is a learning rate. 
	}
	\caption{The algorithms of the reversible network learning with the latent feature generation for a single batch.}
	\label{alg:1}
\end{algorithm}
 
The solution for the above issue using RL is surprisingly simple, and we only need an one steps of feed backward and forward process in Eq. \ref{equ:3} and Eq. \ref{equ:1} in RN. Reversibility of RN can apply to generate latent features, by providing reverse mapping from the likelihood of class to latent feature. The hard sample is considered as unrecognizable samples using a model under the close-set condition, and it is represented as follow: $i_{a}\neq\text{argmax}_{i}p(x|\theta_{i})$, where $x$ and $i_{a}$ denote a hard sample and the annotated class $a$, $\theta_{i}$ represents the parameters for probability model of $i$ class. $p(x|\theta_{i})$ is the likelihood corresponding to class $i$. 

In feed-backward process of RN, generating the latent features can be interpreted by $p^{-1}(\hat{o}|\theta) = \bar{\alpha}$, where $\hat{o}$ is given likelihood data for generating a corresponding latent feature, and $\bar{\alpha}$ is the generated latent feature. Above process can be applied to generate the latent features corresponding to hard samples. This process to generate the latent features corresponding to hard samples can be represented with Eq. \ref{equ:4} as follows:
\begin{equation}
\label{equ:7}
\begin{aligned}
\gamma(\text{tr}(o)\cdot{}W_{t}^{T}+b_{t-1})=\hat{\alpha},
\end{aligned}
\end{equation}
where $\text{tr}$ is a transformation function for an output to modify the likelihood value on output $o$. In this work, we select some elements among the elements in an output vector randomly and assign a value which is similar to the maximum likelihood. The detail method to modifying the $o$ and applying to network training are described at Algorithm. \ref{alg:1}. 
A further process is straightforward. The generated latent features are directly applied to the feed-forward process, and it is equivalent to the general process for image classification. Figure \ref{fig:3:a} illustrates that the process of the latent feature generation on RN, and generation and visualization results of the latent features. 
s

\section{Experiment}
\subsection{Experimental setting and datasets}
We have compared the model applying the reversible manner and the normally trained models. We have implemented a baseline neural network, very deep neural network (VGGnet) \cite{simonyan2014very}, residual network (ResNet) \cite{he2016deep}, and the densely connected convolutional neural network (DenseNet) \cite{huang2017densely}. The structural details of the baseline neural network are shown in table \ref{tbl:1}. In implementing the others, we have employed the structures of VGG-19, ResNet-18, and DenseNet-40 on their studies. Our work is concentrated to demonstrate the efficiency of RL, and not on encourage state-of-the-art performance. Therefore, the experiment is conducted based on the several baseline models intentionally and focused on the comparison between normally trained model and trained model using the reversible learning manner. In our experiments, All networks are trained using stochastic gradient descent (SGD). we employed learning rate decay of 0.0001 and momentum of 0.9. The learning rate is initially set to 0.1, and divided by 10 in 20, 40, and 60 epochs. We conduct a simple data augmentation by cropping and flipping given images. The training and evaluation using each dataset are performed 10 times.  The average values for all experiments are considered as the final quantitative results for each model. All experiments are conducted using Nvidia Titan Xp GPU and 3.20$Ghz$ CPU. The source codes for these experiments are implemented based on Pytorch library.

We demonstrate the efficiencies of RNL through the image classification setting. Af first, we evaluate the models using Cifar-10 and Cifar-100 datasets \cite{krizhevsky2009learning}. The Cifar-10 dataset is composed of 50000 training images and 10000 test images, which can be classified into 10 categories. Each category contains 6000 images. The Cifar-1000 dataset consists of 100 image categories, and each category has 500 training images and 100 test images. The resolution of an image on the dataset is 32 $\times$ 32. When we train the models mentioned above using Cifar-10 and Cifar-100 dataset, we take 128 of batch size in the training stage and 100 batch size in the test stage. All images in Cifar-10 and Cifar-100 datasets are normalized by dividing the channel-wise expectation values when they are inputted to the networks.

 \begin{figure}
        \centering
        \begin{subfigure}{0.22\textwidth}
            \includegraphics[width=\textwidth]{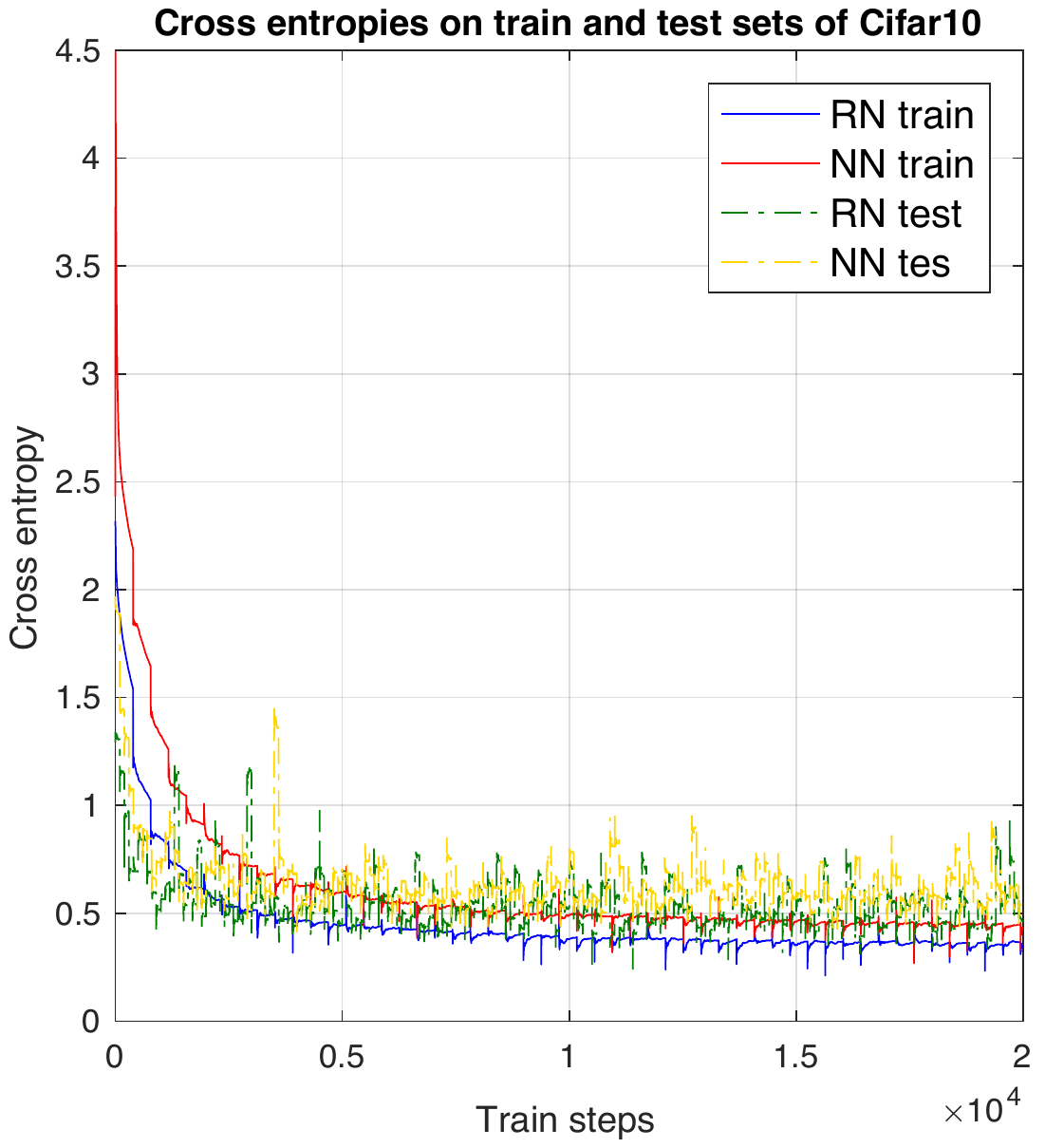}
	\caption{}
	\label{fig:4:a}
        \end{subfigure}
        \hfill
        \begin{subfigure}{0.22\textwidth}
            \includegraphics[width=\textwidth]{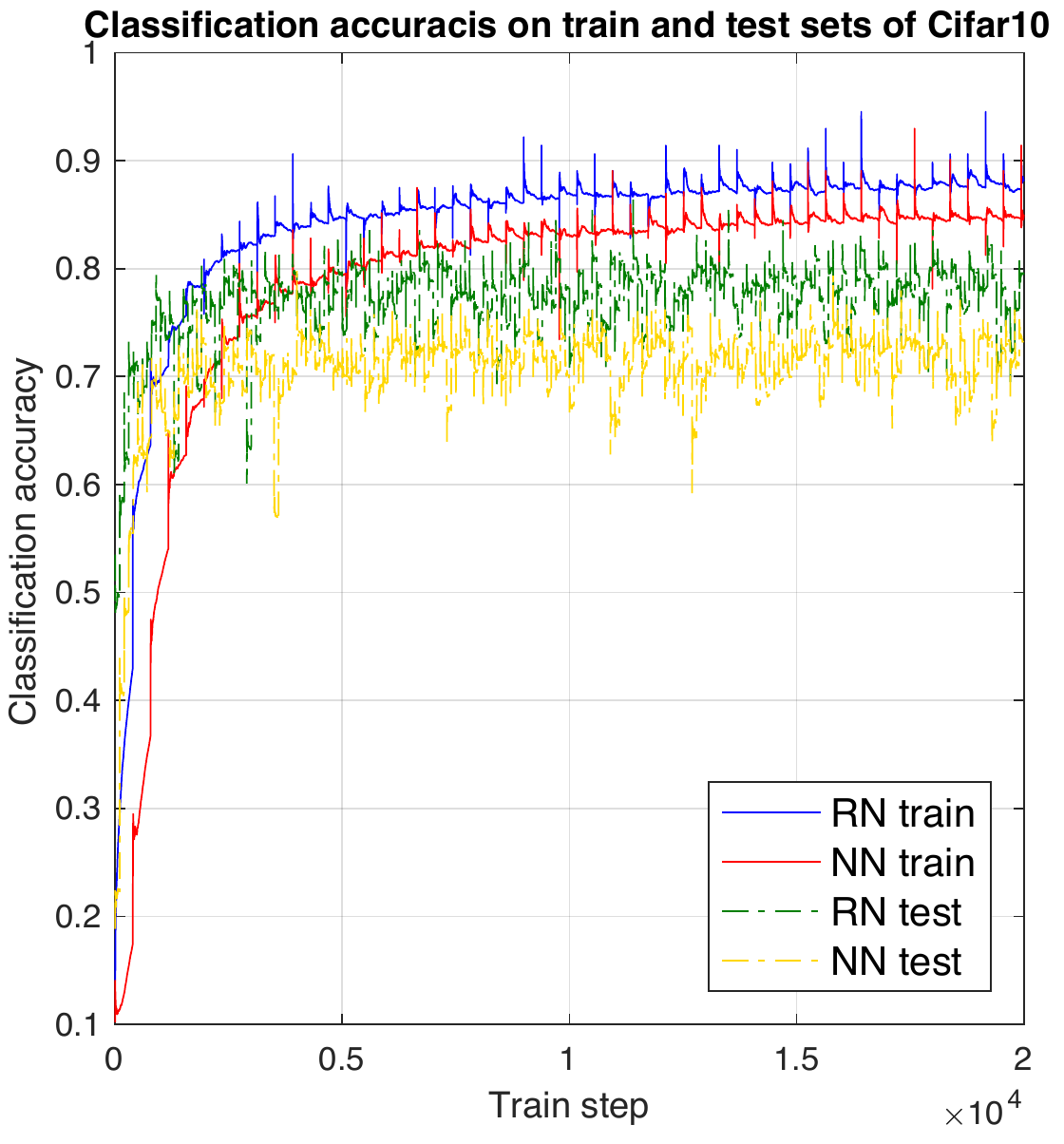}
	\caption{}
	 \label{fig:4:b}
        \end{subfigure}
        \hfill
        \begin{subfigure}{0.22\textwidth}
            \includegraphics[width=\textwidth]{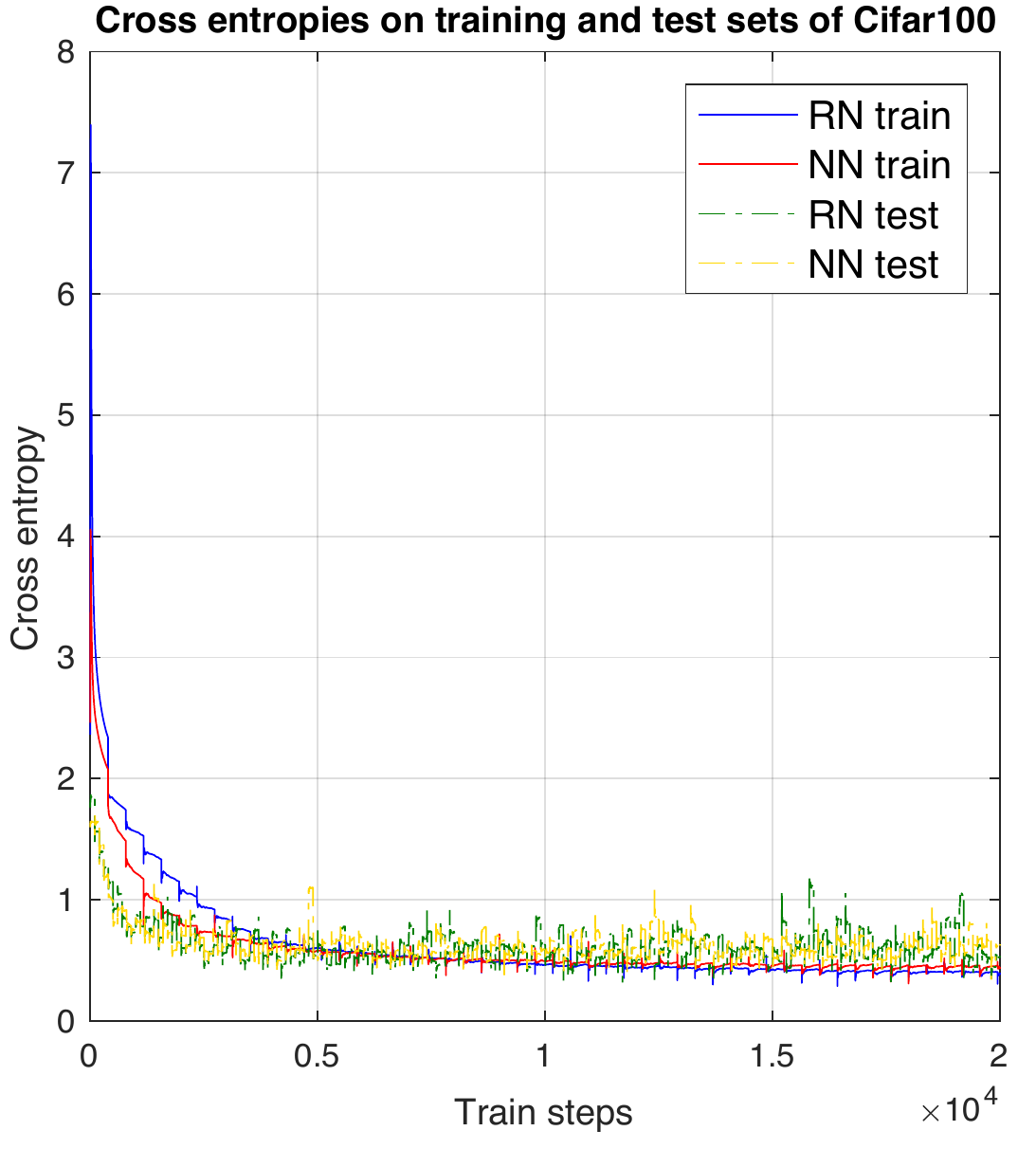}
	\caption{}
	\label{fig:4:c}
        \end{subfigure}
        \hfill
        \begin{subfigure}{0.22\textwidth}
            \includegraphics[width=\textwidth]{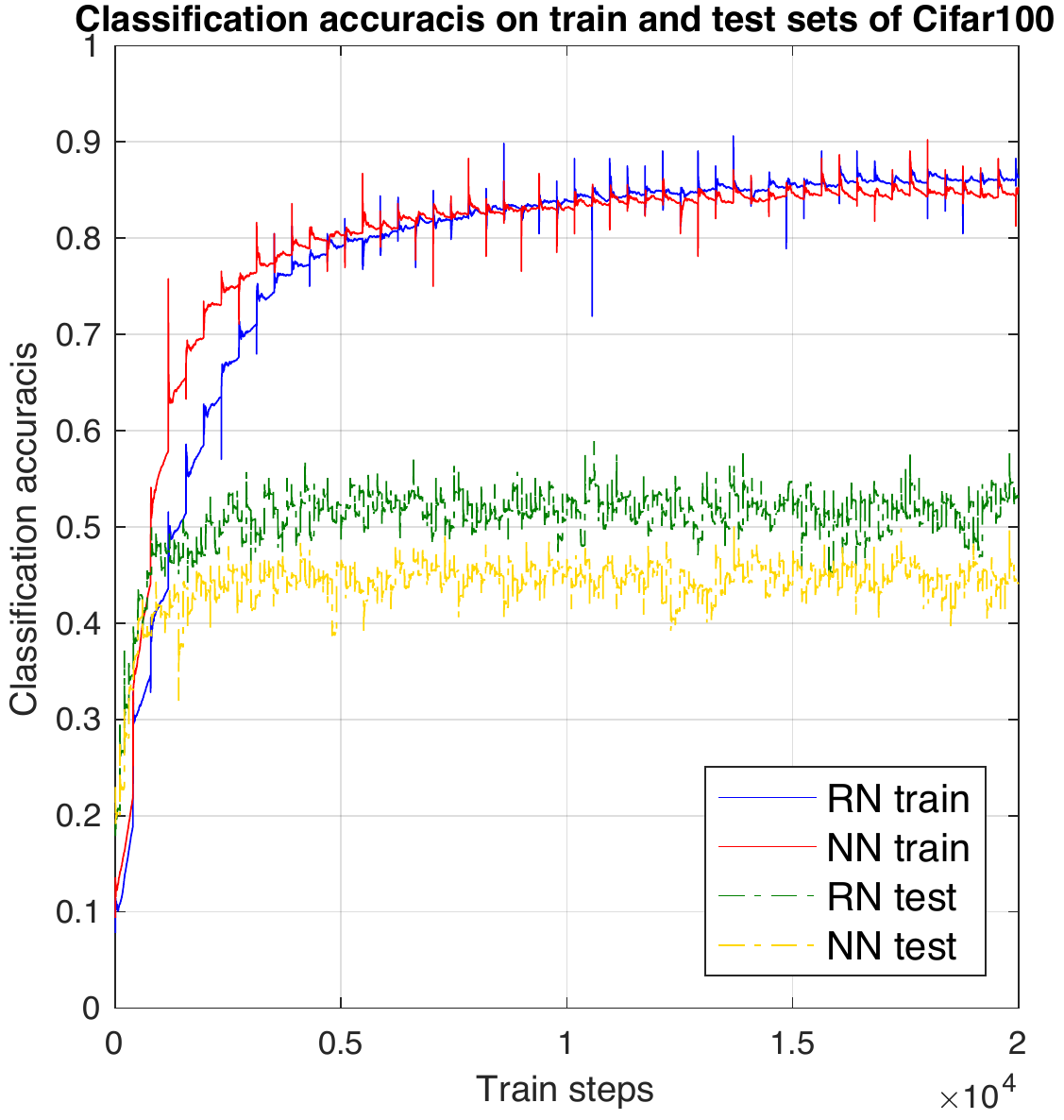}
	\caption{}
	\label{fig:4:d}
        \end{subfigure}
        \hfill
        \caption{(a) and (b) shows the trend of loss and accuracies on training and test sets on Cifar10 dataset respectively. (c) and (d) represents the trend of loss and accuracies on training and test set of Cifar100 dataset respectively. The baseline neural network (NN) and reversible network (RN) are used for this experiment. Solid lines denote that a training set is applied to evaluate models, and dotted lines represented that a test set is used to evaluate models.}
\label{fig:4}  
\vspace{-1ex}
\end{figure}

\begin{table}
    \begin{minipage}{.33\textwidth}
      \centering
       \resizebox{\textwidth}{!}{
        \begin{tabular}{l|c|c}
        \hline
        Layer & Kernel & Act \\ \hline\hline
        Conv &  5$\times{}$5$\times$3$\times$32 & L-Relu \\\hline
        Conv &  5$\times{}$5$\times$32$\times$32 & L-Relu \\\hline
        Max-pool & - & - \\\hline
        Conv & 5$\times{}$5$\times$32$\times$64 &  L-Relu \\\hline
        Conv & 5$\times{}$5$\times$64$\times$64 & L-Relu \\\hline
        Max-pool & - & -\\\hline
        Conv & 5$\times{}$5$\times$64$\times$128 & L-Relu \\\hline
        Conv & 5$\times{}$5$\times$128$\times$128 & L-Relu \\\hline
        Fc1 & 2048$\times{}$256 & L-Relu\\\hline
        Fc2 & 256 $\times{}N$ & Softmax \\\hline
        \end{tabular}
        }
        \caption{Structural detail of baseline neural network applied to the baseline neural network (NN) and the reversible network (RN) on our experiments. $N$ is the number of classes corresponding to a given dataset.}
        \label{tbl:1}
    \end{minipage}
     \hspace{0.5em}
    \begin{minipage}{.64\textwidth}
      \centering
        \resizebox{\textwidth}{!}{
        \begin{tabular}{l|c|cc|cc}
        \hline
        Method & Params & C10 & C10+ & C100 & C100+ \\ \hline\hline
        Baseline-NN &  1.3M & 20.18 & 17.17 & 49.72 & 40.16  \\
        Baseline-RN & 1.3M & 13.62 & 9.17 & 45.19 & 34.61 \\\hline
        VGG-19 \cite{simonyan2014very} &  13.4M & 8.48 & 7.82 & 43.80 & 28.96  \\
        \textit{Reversible}-VGG-19 &  13.4M & 7.12 & 6.94 & 37.56 & 24.91 \\\hline
        ResNet \cite{he2016deep} & 1.7M & 7.92 & 6.53 & 33.41 & 23.24  \\
        \textit{Reversible}-ResNet & 1.7M & 6.01 & 5.94 & 27.72 & 20.03 \\\hline
        DenseNet $(k=12)$\cite{huang2017densely} &  1.0M & 8.01 & 6.47 & 28.15 & 23.24   \\
        \textit{Reversible}-DenseNet & 1.0M & 5.84 & \textbf{5.17} & 22.17 & \textbf{20.94}\\\hline
        \end{tabular}%
        }
        \caption{Error rates (\%) on Cifar-10 and Cifar-100. '\textit{Reversible}' denotes the model is trained with the proposed reconstruction error. $+$ indicates that the data augmentation based on simple image transformation is used. The bolded value is the best performance in our experiments.}
        \label{tbl:2}
    \end{minipage} 
\vspace{-1ex}
\end{table}

In addition to the experiments using Cifar-10 and Cifar-100 datasets. We have carried out additional experiments with an Extremely Biased and Poorly Categorized (EBPC) dataset\footnote{The EBPC dataset is available at \url{https://github.com/andreYoo/CED-algorithm}}. We propose EBPC dataset for observing a network performance when networks are trained with a highly unbalanced and terribly classified dataset. EBPC dataset is constructed by combining several public datasets roughly, and the dataset has 3,470 classes and consists of 271,516 images for the training and 82,771 images for the test. The datasets which are used to construct EBPC dataset have proposed for image classification, face recognition, and person re-identification. The datasets used to construct the EBPC dataset as follows: 1) MNIST dataset \cite{lecun1998gradient}, 2) Cifar-10 \& 100 datasets \cite{krizhevsky2009learning}, 3) Stanford dog dataset \cite{khosla2011novel}, 4) Flowers dataset with 101 categories \cite{nilsback2008automated}, LFW Face dataset \cite{learned2016labeled}, and CUHK03 dataset \cite{li2014deepreid}. Even if several labels take homogeneous, these are identified as different classes in the EBPC dataset. For example, the \textit{automobile} class in Cifar-10 dataset and the \textit{vehicles} class in Cifar-100 dataset are considered as different classes. We did not increase the number of samples in each class artificially, and we only normalized the image size of each dataset as $64\times{}64$.  EBPC dataset consists of 3470 class, and each class has a minimum 2 and maximum 10000 samples. The details of each dataset and the quantitative properties of EBPC dataset are shown in table \ref{tbl:3}. As same as the experiments using Cifar-10 and Cifar-100 datasets, we set 128 batch size and 100 batch size for the training and test the network models respectively. 

 \begin{figure}
        \centering
        \begin{subfigure}{0.49\textwidth}
            \includegraphics[width=\textwidth]{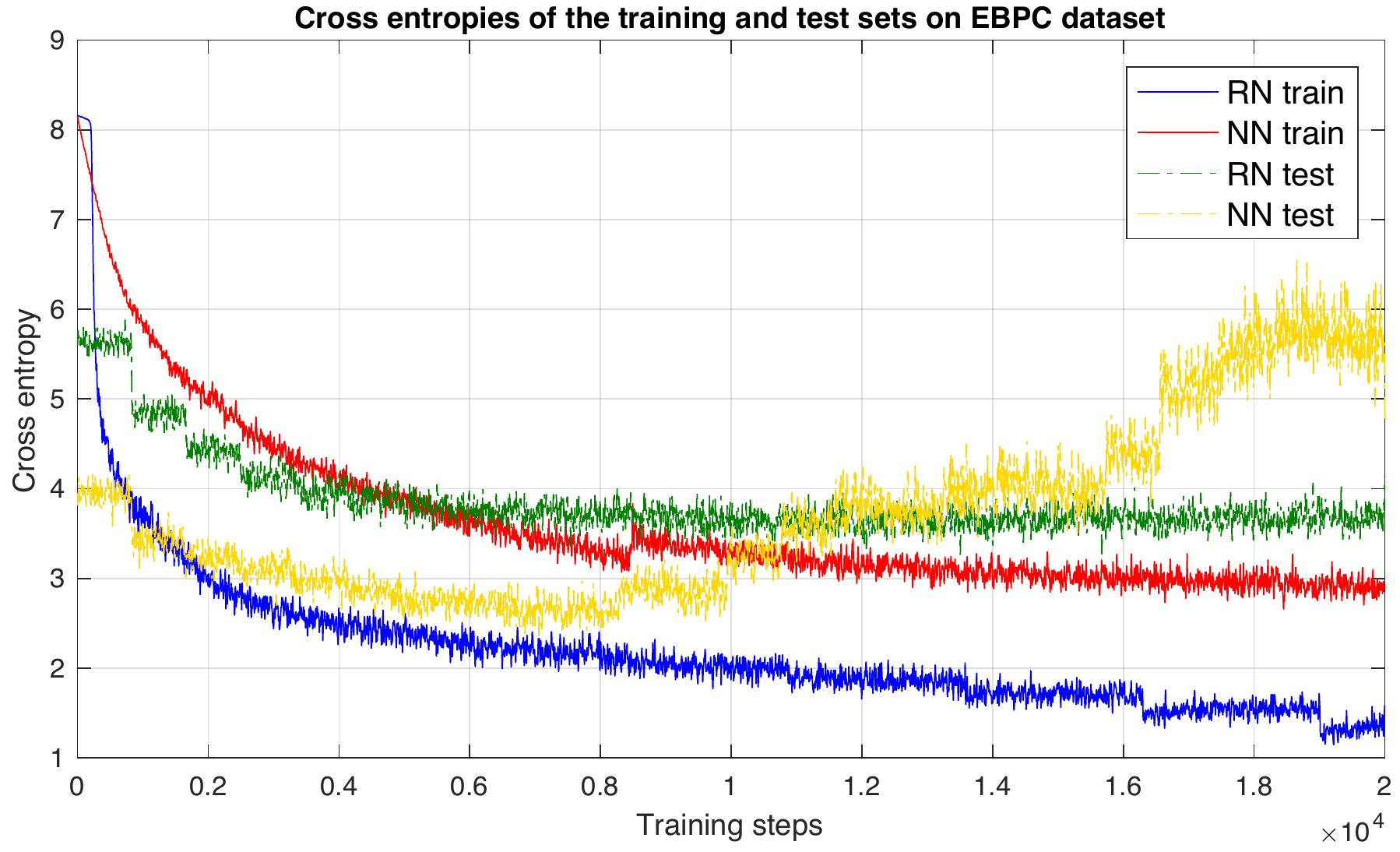}
	\caption{}
	\label{fig:5:a}
        \end{subfigure}
        \hfill
        \begin{subfigure}{0.49\textwidth}
            \includegraphics[width=\textwidth]{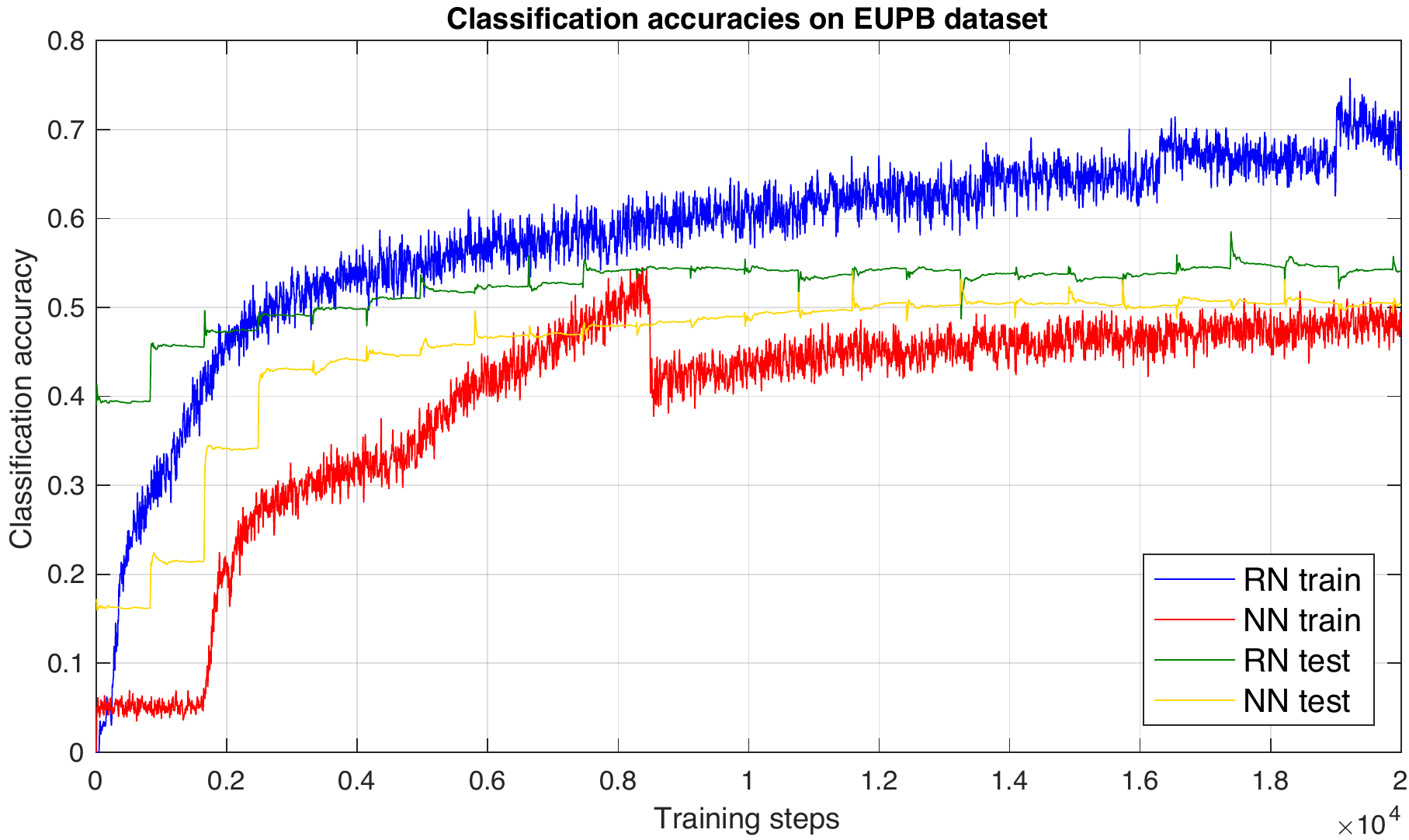}
	\caption{}
	 \label{fig:5:b}
        \end{subfigure}
        \hfill
        \caption{(a) shows the trend of loss on training and test sets on EBPC dataset respectively. (b) shows the trend of accuracies on training and test sets on EBPC dataset respectively. Solid lines denote that a training set is applied to evaluate models, and dotted lines represented that a test set is used to evaluate models.}
\label{fig:5}
\end{figure}
\begin{table}
    \begin{minipage}{.54\textwidth}
      \centering
       \resizebox{\textwidth}{!}{
        \begin{tabular}{l|c|c|c|c|c}
        \hline
        Dataset & Subject & Class\# & Train\# & Test\# & $\textit{SPC}_{t}$ \\ \hline\hline
        Cifar-10 \cite{krizhevsky2009learning} & IC & 10 & 50000 & 10000 & 5000\\\hline
        Cifar-100 \cite{krizhevsky2009learning} & IC & 100 & 50000 & 10000 & 500 \\\hline
        Flower102 \cite{nilsback2008automated} & IC & 102  & 1020 & 6149 & 102 \\\hline
        CUHK03 \cite{li2014deepreid} & PRI & 1467 & 19574 & 8619 & 13.3 \\\hline
        LFW \cite{huang2008labeled} & FR & 1650 & 5665 & 3391 & 3.4 \\\hline
        MNIST \cite{lecun1998gradient} & IC & 10 & 60000 & 10000 & 6000 \\\hline
        Stanford \cite{khosla2011novel} & IC & 120 & 12000 & 8580 &1000 \\\hline
        SVNH \cite{netzer2011reading} & IC & 10 & 73257 & 26032 & 7325.7 \\\hline
        Total & - & 3470  & 271516 & 82771 & - \\\hline
        \end{tabular}
        }
        \caption{Composition of the extremely biased and poorly categorized (EBPC) dataset. $\textit{SPC}_{t}$ denotes the number of 'samples per class' on train set of each dataset, and it is computed by $\frac{Train\#}{Class}$. 'IC', 'PRI', and 'FR' denote 'Image classification', 'Person re-identification', and 'Face recognition'}
        \label{tbl:3}
    \end{minipage}%
     \hspace{1em}
    \begin{minipage}{.43\textwidth}
      \centering
        \resizebox{\textwidth}{!}{
        \begin{tabular}{l|c|cc}
        \hline
        Method &  EBPC & EBPC+ \\ \hline\hline
        Baseline-NN & 50.27  & 45.83  \\
        Baseline-RN & 44.19  & 36.88  \\\hline
        VGG-16 \cite{simonyan2014very} &  39.97 & 36.71 \\
        \textit{Reversible}-VGG-16  & 34.18 & 31.59 \\\hline
        ResNet-18 \cite{he2016deep} & 39.74 & 34.64  \\
        \textit{Reversible}-ResNet-18 & 32.36 & 30.86 \\\hline
        DenseNet-32 \cite{huang2017densely} & 37.51 & 31.95  \\
        \textit{Reversible}-DenseNet-32 & \textbf{31.88} & \textbf{28.74} \\\hline
        \end{tabular}%
        }
        \caption{Error rates (\%) on EBPC dataset. '\textit{Reversible}' denotes the model is trained with the proposed reconstruction error. $+$ indicates that simple data augmentation is used. The bolded value is the best performance in our experiments.}
        \label{tbl:4}
    \end{minipage}
    \vspace{-1ex}
\end{table}

\subsection{Quantitative comparison}
Table \ref{tbl:2} contains the classification error of listed network models on Cifar-10 and Cifar-100 datasets. The model achieving the lowest classification error is the reversible-DenseNet applying a simple data augmentation. The model shows 5.17\% of classification error on Cifar-10 dataset and 20.94\% of classification error on Cifar-100 dataset. Among the experimental results using ResNet, the lowest errors for Cifar-10 and Cifar-100 datasets are 5.94\% and 20.03\% respectively. In experimental results using VGG-19, 6.94\% and 24.91\% are the lowest classification errors on Cifar-10 and Cifar-100 datasets. The experimental results show clear advantages over current deep neural network models and a lot of compared baselines. The models trained with the proposed reversible learning achieve better classification errors than the others. In the experimental results of baseline models, the network trained with the proposed method shows at least 8\% better classification errors whether the simple data augmentation is applied or not. The evaluation results using other network models show a similar trend to the experiment using the baseline network. 

In experimental results using EBPC dataset, the lowest classification error is 28.74\%, and this figure has achieved by the DenseNet trained with the reversible learning manner, and the data augmentation. The ResNet model, which respectively achieved 5.94\% and 20.03\% classification errors on Cifar-10 and Cifar-100 datasets, recorded 40.08\% error on the experiment using EBPC dataset. VGG-19 also achieve 41.59\% of classification error in the experiment. The overall classification performances evaluated using EBPC dataset are lower than the performances on Cifar-10 and Cifar-100 datasets. The typically trained DenseNet achieves 31.88\% classification errors, and this figure is 3.14\% larger than the reversibly trained model. As same as the experimental results using DenseNet, the experimental results using VGG-19 and ResNet also shows a similar trend to the experimental results using DenseNet. In evaluation results using VGG-19, the VGG-19 trained with RNL shows 3\% lower classification errors than the others. The experimental results using ResNet also shows the ResNet trained by RNL achieves better performance than the other. The classification accuracies using EBPC dataset are presented in table \ref{tbl:4}.

\subsection{Analysis}
The experimental results show clear advantages over current deep neural network models and a lot of compared baselines. The experimental results show that the network model trained with RNL outperformed the normally trained models. The most noticeable things in our experiment are that the models trained to improve reversibility of networks achieve better performance whether the performance differences are small or large collectively.

Our interpretation of these performance improvements is as follows. As we mentioned in Section 2, the latent feature generation method based on RNL can influence recognition performance in a model based on the neural network. We tried to improve mapping functionality using the RNL. The RNL can encourage the reversibility of neural networks, which can reconstruct input data on supervised learning setting. In the learning procedure, the proposed RNL plays a critical role to improve the network reversibility explicitly. The experimental results on Cifar-10 and Cifar-100 shows the model trained by RNL achieve better classification errors than the others. Not only classification errors, but also the descending trends of loss also shows that the models applying RNL achieves better performances. Figure \ref{fig:4:a} and figure \ref{fig:4:c} represent the loss trend graphs of baseline network models, which are trained with RNL and general training process, using Cifar-10 and Cifar-100 datasets. The models are trained and tested by the training sets and test sets on Cifar-10 and Cifar-100 datasets. These graphs show that the networks applying RNL take the lower loss than the others. Additionally, the graphs for classification accuracy trend during network training, which are shown in figure \ref{fig:4:b} and figure \ref{fig:4:d} also show similar circumstance. 

Not only the experimental results on Cifar-10 and Cifar-100 dataset but also the experimental results on EBPC dataset also shows that the models improving the reversibility can achieve better classification accuracies than the others that trained normally. Figure \ref{fig:5} illustrates that the trend of loss and accuracy of the baseline network model depending on training manners. Both the cross-entropy graphs and the classification accuracy graphs present that the networks trained with RNL can provide better classification performance than others. Interestingly, in contrast to the cross-entropy curve of RN on the test set of EBPC dataset is gradually decreased during training, the curve of the cross-entropy of NN represents that the cross-entropy increases during training. It may mean that the RNL using latent feature generation can be considered as a stable learning method when a biased and poorly classified dataset is given.

\section{Conclusion}
In this paper, we propose the reversible learning method to boost the mapping capability of neural networks. The proposed method generates and learns the latent features regarded as samples which have a lower likelihood automatically. Thus, it can improve the mapping capability of neural networks without both additional data augmentation and a complementary process for resampling a given dataset accordingly. Also, it can be a memory and cost-effective approach since it is not a method for augmenting or generating samples for dataset itself and generates latent features which have lower dimensionality than given samples. Additionally, the proposed method does not require modification on network structures or loss functions, and it may be easily applied to the various recognition methods using neural networks, not only visual recognition but also for speech recognition. The experimental results show that the network models trained with the proposed method can outcome the performance of existing models. 
\small
\bibliographystyle{unsrt}
\bibliography{egbib}

\begin{thebibliography}{10}

\bibitem{lecun1998gradient}
Yann LeCun, L{\'e}on Bottou, Yoshua Bengio, and Patrick Haffner.
\newblock Gradient-based learning applied to document recognition.
\newblock {\em Proceedings of the IEEE}, 86(11):2278--2324, 1998.

\bibitem{he2016deep}
Kaiming He, Xiangyu Zhang, Shaoqing Ren, and Jian Sun.
\newblock Deep residual learning for image recognition.
\newblock In {\em Proceedings of the IEEE conference on computer vision and
  pattern recognition}, pages 770--778, 2016.

\bibitem{wen2016discriminative}
Yandong Wen, Kaipeng Zhang, Zhifeng Li, and Yu~Qiao.
\newblock A discriminative feature learning approach for deep face recognition.
\newblock In {\em European Conference on Computer Vision}, pages 499--515.
  Springer, 2016.

\bibitem{liu2017sphereface}
Weiyang Liu, Yandong Wen, Zhiding Yu, Ming Li, Bhiksha Raj, and Le~Song.
\newblock Sphereface: Deep hypersphere embedding for face recognition.
\newblock In {\em The IEEE Conference on Computer Vision and Pattern
  Recognition (CVPR)}, volume~1, page~1, 2017.

\bibitem{huang2017densely}
Gao Huang, Zhuang Liu, Laurens Van Der~Maaten, and Kilian~Q Weinberger.
\newblock Densely connected convolutional networks.
\newblock In {\em CVPR}, volume~1, page~3, 2017.

\bibitem{viola2001robust}
Paul Viola, Michael Jones, et~al.
\newblock Robust real-time object detection.
\newblock {\em International journal of computer vision}, 4(34-47):4, 2001.

\bibitem{dollar2014fast}
Piotr Doll{\'a}r, Ron Appel, Serge Belongie, and Pietro Perona.
\newblock Fast feature pyramids for object detection.
\newblock {\em IEEE transactions on pattern analysis and machine intelligence},
  36(8):1532--1545, 2014.

\bibitem{dong2018imbalanced}
Qi~Dong, Shaogang Gong, and Xiatian Zhu.
\newblock Imbalanced deep learning by minority class incremental rectification.
\newblock {\em IEEE transactions on pattern analysis and machine intelligence},
  2018.

\bibitem{chawla2002smote}
Nitesh~V Chawla, Kevin~W Bowyer, Lawrence~O Hall, and W~Philip Kegelmeyer.
\newblock Smote: synthetic minority over-sampling technique.
\newblock {\em Journal of artificial intelligence research}, 16:321--357, 2002.

\bibitem{kumar2010self}
M~Pawan Kumar, Benjamin Packer, and Daphne Koller.
\newblock Self-paced learning for latent variable models.
\newblock In {\em Advances in Neural Information Processing Systems}, pages
  1189--1197, 2010.

\bibitem{chang2017active}
Haw-Shiuan Chang, Erik Learned-Miller, and Andrew McCallum.
\newblock Active bias: Training more accurate neural networks by emphasizing
  high variance samples.
\newblock In {\em Advances in Neural Information Processing Systems}, pages
  1002--1012, 2017.

\bibitem{jiang2017mentornet}
Lu~Jiang, Zhengyuan Zhou, Thomas Leung, Li-Jia Li, and Li~Fei-Fei.
\newblock Mentornet: Regularizing very deep neural networks on corrupted
  labels.
\newblock {\em arXiv preprint arXiv:1712.05055}, 4, 2017.

\bibitem{schroff2015facenet}
Florian Schroff, Dmitry Kalenichenko, and James Philbin.
\newblock Facenet: A unified embedding for face recognition and clustering.
\newblock In {\em Proceedings of the IEEE conference on computer vision and
  pattern recognition}, pages 815--823, 2015.

\bibitem{wu2017sampling}
Chao-Yuan Wu, R~Manmatha, Alexander~J Smola, and Philipp Krahenbuhl.
\newblock Sampling matters in deep embedding learning.
\newblock In {\em Proceedings of the IEEE International Conference on Computer
  Vision}, pages 2840--2848, 2017.

\bibitem{ren2018learning}
Mengye Ren, Wenyuan Zeng, Bin Yang, and Raquel Urtasun.
\newblock Learning to reweight examples for robust deep learning.
\newblock {\em arXiv preprint arXiv:1803.09050}, 2018.

\bibitem{xu2014deep}
Li~Xu, Jimmy~SJ Ren, Ce~Liu, and Jiaya Jia.
\newblock Deep convolutional neural network for image deconvolution.
\newblock In {\em Advances in Neural Information Processing Systems}, pages
  1790--1798, 2014.

\bibitem{simonyan2014very}
Karen Simonyan and Andrew Zisserman.
\newblock Very deep convolutional networks for large-scale image recognition.
\newblock {\em arXiv preprint arXiv:1409.1556}, 2014.

\bibitem{krizhevsky2009learning}
Alex Krizhevsky and Geoffrey Hinton.
\newblock Learning multiple layers of features from tiny images.
\newblock Technical report, Citeseer, 2009.

\bibitem{khosla2011novel}
Aditya Khosla, Nityananda Jayadevaprakash, Bangpeng Yao, and Fei-Fei Li.
\newblock Novel dataset for fine-grained image categorization: Stanford dogs.
\newblock In {\em Proc. CVPR Workshop on Fine-Grained Visual Categorization
  (FGVC)}, volume~2, 2011.

\bibitem{nilsback2008automated}
Maria-Elena Nilsback and Andrew Zisserman.
\newblock Automated flower classification over a large number of classes.
\newblock In {\em 2008 Sixth Indian Conference on Computer Vision, Graphics \&
  Image Processing}, pages 722--729. IEEE, 2008.

\bibitem{learned2016labeled}
Erik Learned-Miller, Gary~B Huang, Aruni RoyChowdhury, Haoxiang Li, and Gang
  Hua.
\newblock Labeled faces in the wild: A survey.
\newblock In {\em Advances in face detection and facial image analysis}, pages
  189--248. Springer, 2016.

\bibitem{li2014deepreid}
Wei Li, Rui Zhao, Tong Xiao, and Xiaogang Wang.
\newblock Deepreid: Deep filter pairing neural network for person
  re-identification.
\newblock In {\em CVPR}, 2014.

\bibitem{huang2008labeled}
Gary~B Huang, Marwan Mattar, Tamara Berg, and Eric Learned-Miller.
\newblock Labeled faces in the wild: A database forstudying face recognition in
  unconstrained environments.
\newblock In {\em Workshop on faces in'Real-Life'Images: detection, alignment,
  and recognition}, 2008.

\bibitem{netzer2011reading}
Yuval Netzer, Tao Wang, Adam Coates, Alessandro Bissacco, Bo~Wu, and Andrew~Y
  Ng.
\newblock Reading digits in natural images with unsupervised feature learning.
\newblock 2011.

\end{thebibliography}
\end{document}